\documentclass[10pt]{article}

\usepackage[margin=0.9in]{geometry}
\usepackage[figuresright]{rotating}

\usepackage{graphicx,color}
\usepackage{amsmath}
\usepackage{multirow}

\usepackage[ruled,vlined]{algorithm2e}

\title{Predictive User Modeling with Actionable Attributes\\ \vspace{2 mm} {\large Technical Report}}

\author{
  Indr\.e \v{Z}liobait\.e\thanks{This research was done in 2010-2011 when I.~\v{Z}liobait\.e was affiliated with Eindhoven University of Technology, The Netherlands.}\\
  Aalto University and HIIT, Finland\\
  \texttt{indre.zliobaite@aalto.fi}
  \and
  Mykola Pechenizkiy\\
  Eindhoven University of Technology, The Netherlands\\
  \texttt{m.pechenizkiy@tue.nl}
}

\date{}

\begin{document}
\maketitle

\begin{abstract}
Different machine learning techniques have been proposed and used for modeling individual and group user needs, interests and preferences.
In the traditional predictive modeling instances are described by observable variables, called attributes.
The goal is to learn a model for predicting the target variable for unseen instances.
For example, for marketing purposes a company consider profiling a new user based on her observed web browsing behavior, referral keywords or other relevant information.
In many real world applications the values of some attributes are not only observable, but can be actively decided by a decision maker.
Furthermore, in some of such applications the decision maker is interested not only to generate accurate predictions, but to maximize the probability of the desired outcome.
For example, a direct marketing manager can choose which type of a special offer to send to a client (actionable attribute), hoping that the right choice will result in a positive response with a higher probability.
We study how to learn to choose the value of an actionable attribute in order to maximize the probability of a desired outcome in predictive modeling.
We emphasize that not all instances are equally sensitive to changes in actions.
Accurate choice of an action is critical for those instances, which are on the borderline (e.g. users who do not have a strong opinion one way or the other).
We formulate three supervised learning approaches for learning to select the value of an actionable attribute at an instance level.
We also introduce a focused training procedure which puts more emphasis on the situations where varying the action is the most likely to take the effect.
The proof of concept experimental validation on two real-world case studies in web analytics and e-learning domains highlights the potential of the proposed approaches.
\end{abstract}

\section{Introduction}

Over the past decade the use of machine learning and data mining for user modeling have been studied in different e-commerce, e-learning and other domains generating vast amount of user-related data including the click-steam, usage logs, social media entries, event logs, and many other user behavior traces captured by (web-based) information systems.

User modeling poses a number of challenges for predictive machine learning and statistical modeling,
including but not limited to the availability of large enough and good quality (labeled) data to train the models, temporal dynamics, concept drift, and computational complexity~\cite{Webb:2001:MLforUM}.
Many of these challenges have been addressed successfully by attempting to adopt existing and to develop new supervised learning approaches for building real adaptive and personalized information systems, heavily relying on data-driven user modeling and decision making.

Some of the success stories of predictive user modeling are widely known.
They include, for instance, new approaches for dealing with temporal dynamics in collaborative filtering recommender systems~\cite{Koren10}, adoption and assembling of existing approaches for building effective content-based filtering systems, e.g. for adaptive news access~\cite{adaptivenews00}.
However, e-commerce and recommenders in particular is not the only area where machine learning shows promising results.
For instance, predicting student drop out from the university~\cite{DekkerEDM09} or guiding students through the enrollment process~\cite{DBLP:journals/umuai/SacinCPAVEO11} based on academic performance are illustrative example of user (student) modeling in the traditional educational domain.
Another recent study reports on modeling consumer choice behavior in selecting long distance communication modes over time \cite{Hu08}.
Nevertheless, a recent market survey calls for research to focus on predictive models applicable for internet marketing, long-term effects of direct marketing, irritation from direct marketing offers, and segmentation and predictive modeling techniques \cite{Verhoef03}.

A wide variety of real world applications, in which effective predictive user modeling is essential, facilitates continuous extension of machine learning tool sets for predictive user modeling, as well as continuous incremental improvement of the performance of particular techniques.
Furthermore, better understanding of needs and possibilities for user modeling in different real applications triggers the consideration of new problem formulations and peculiar settings, in which predictive models are being learnt. This paper presents one of such new considerations.


In supervised learning employed for predictive modeling it is typically assumed that the values of attributes are observed and the task is to predict the values of the target variable (e.g. class label). However, in a number of predictive analytics applications the attribute space consists not only of observable but also of actionable attributes, the values of which can be pro-actively decided by decision makers.
Consider a direct marketing task: given an individual to be provided with a piece of information about a cable TV subscription, the task is to predict how likely she is to subscribe it.
Given a historical dataset with known ground truth (which individuals responded positively, which did not), the task is
to provide a model for supporting business decisions to which users what offers to send, and to whom not to send any offer at all.

In the considered case, the values of some attributes can be chosen (manipulated) by a decision maker.
We call such attributes \emph{actionable}.
For a toy example, a marketer may choose whether to send an offer in a blue or yellow envelope when thinking how to maximize the probability of the desired outcome (in this case a subscription).

This study focuses on \emph{learning models for selecting} the value of the actionable attribute so that the probability of a desired outcome is maximized.
In the direct marketing example, given an individual, we learn to choose, which envelope to use (that is the actionable attribute), in order to maximize the probability of a positive response (the target variable).

\begin{figure}[h]
\centering
\includegraphics[width=3in]{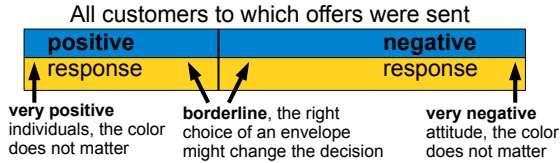}
\caption{Illustration of the setting, a marketing example.}
\label{fig:illus}
\end{figure}

The choice is evident if one action is obviously better than the other (e.g. discount is better than no discount), but the reality is more complex than that.
Our study is motivated by an observation, that different users may be sensitive to actions in different ways.
First, some users may prefer (i.e.\ be sensitive to or responsive to) one action, others may prefer the other.
In addition, some users will certainly not accept the offer, and others will certainly accept it no matter which envelope is used.
Yet, there will be a group of users whose final decision may be affected by one or the other color of the envelope.
Thus, we consider the setting where for \emph{some} users yellow, and for \emph{some} users blue color is more attractive, while \emph{others} do not care,
as illustrated in Figure~\ref{fig:illus}.

Our task is to learn a decision rule that would decide which of the alternative actions to select for a user in consideration.
This task is not trivial,
since information whether a user has accepted the offer because of the correct action (envelope color) or for some other reasons is typically not available in the historical data.
Thus, one of the major challenges in such a learning task is to formulate a representative learning objective using the available historical data.
Without access to the ground truth such objectives can be expected to be noisy or even partially incorrect.
The good news is that employing some reasoning about the user behavior can help to compensate for such incorrectness.
We can argue that the correct choice of action is critical for those users, who are the most sensitive to actions, as illustrated in Figure \ref{fig:illus}.
Therefore, instead of optimizing the prediction accuracy over all the instance space we can focus model training on the sensitive cases.
Sensitive cases include, but are not limited to the decision boundary between the positive and the negative instances (as shown in the figure).
As a compensation, we can tolerate inaccurate predictions in the situations, where the choice of action does not change the final outcome.

This study presents the following main contributions.
\begin{itemize}
  \item We formulate a new problem setting from the machine learning point of view. We would like to emphasize that the technical contribution of the paper is primarily in formulating the setting and the evaluation framework, as a solution to the problem of business decision support with actionable attributes. In this paper we are solving a different problem than the traditional classification techniques do.
  \item We introduce a general approach and three concrete techniques for supervised learning instance-level action selection rules from offline data. 
  \item We propose the focused training procedure, which allows to optimize the predictive model performance for the most sensitive cases by trading off the performance in situations where the action does not change the final outcome.
  \item We provide an empirical evidence that the considered settings are reasonable and the proposed approaches are needed. The problem we address cannot be reduced to learning a classifier with reject option, because the sensitive cases can be anywhere, not only close to the decision boundary.
      We also show the validity of the proposed approaches on synthetic and real-world datasets.
\end{itemize}
A short version of this study appeared in the proceedings of the workshop on Domain Driven Data Mining~\cite{Zliobaite10DDDM}. 


The paper is organized as follows.
In Section~\ref{sec:background} we define and discuss the problem and assumptions taken in this study.
In Section~\ref{sec:method} we introduce three supervised learning techniques to select the value of an actionable attribute at an instance level.
Section~\ref{sec:literature} discusses related work.
Section~\ref{sec:experiments} highlights the peculiarities of the evaluation, presents the validation results on a synthetic dataset and exploration results on real datasets from two case studies.
In Section~\ref{sec:discussion} we discuss complementary and further research scenarios and challenges.
Section~\ref{sec:conclusion} concludes the study.

\section{Problem setting}
\label{sec:background}


Predictive analytics is often used for business decision support.
The idea is to build a computational model using a historical dataset for making predictions for operating decision for the future.
Typically, a set of input attributes is given and the goal is to predict the outcome (the target label).
Sometimes it may be possible to affect the outcome by management actions.

\emph{Actionable attribute} 
is an attribute, the value of which can be changed by actions of a decision maker prior to predicting the target label.
The values of an \emph{observable attribute} 
cannot be changed.
For example, a type of treatment for a particular patient can be decided by a doctor, while the age of the patient is only observable.
Examples of actionable attributes in predictive analytics:
\begin{itemize}
\item deciding what credit amount to offer when the task is to predict (assess) credit risk (offering a smaller amount often decreases the credit risk);
\item offering complementary products, discounts, communication channel of the offer when the task is to predict the response rate in direct marketing;
\item choosing the type of a recommender technique and the items to show when the task is to predict preferences for recommendation;
\item carrying out promotion activities when the task is to predict sales volume for a given product;
\item deciding upon the bidding amount when the task is to predict the convergence rate in the sponsored search advertisement.
\end{itemize}
Depending on the task one or more attributes can be actionable. For example, in the direct marketing task offering a complementary product and choosing a communication channel are two actionable attributes that could be combined in one task.
To keep the focus in this study we assume that there is only one actionable attribute $a$ and consider that the cost of executing this action is the same for any instance.
We also assume that the data distribution is stationary over time.

In the traditional supervised learning a set of instances with their class labels $({\bf Z},{\bf y})$ is given.
The goal is to learn a model ${\cal L}:Z\rightarrow y$ that maps any $p$-dimensional instance $Z$ to its true label $y$.
The desired performance is high accuracy of predicting $y$ from $Z$ on unseen data.

The problem setting for learning with actionable attributes can be defined as follows.
Let\\ $Z=\{X A\}=\{x^{(1)} x^{(2)} \ldots x^{(m)} a^{(1)} \ldots a^{(k)}\}$ be an instance in $p$-dimensional input space, where there are $m$ observable attributes and $k$ actionable attributes, $m+k=p$ and $y$ is the class label (outcome).
Suppose that there is some value of the label $y$ that is more desired that others; we denote it as $y^\ast$.
The goal is to find a model ${\cal H}:X \rightarrow a$ (action rule) such that $P(y^\ast|X,a)$ is maximized, where $a$ is an actionable attribute and $X$ is an example in the input space without the actionable attribute.
Here the desired performance is maximizing the number of events $y^\ast$ on unseen data.

The task is difficult, as ideally for estimating the effects of actions one would need a dataset recording the effects of different actions to the same user under the same conditions.
However, real-world user data is highly unlikely to include pairs of the identical twins to which different actions have been applied.
Typically, in business data actions have been applied to individuals either at random or using some fully or partially deterministic procedure.
Therefore, it is not certain what would have happen in case of an alternative action.

At best, for collecting such kind of training data one could employ an A/B or the multivariate testing procedure\footnote{A/B test refers to a direct comparison between two design alternatives in a controlled experiment, i.e. everything in the system is kept unchanged except the item under test, which is switched between the two alternatives; see e.g.~\cite{Crook:2009:SPA:1557019.1557139} for the discussion of the state-of-the-art in controlled experimentation for e-commerce applications development.}.
In marketing A/B testing is realized by so called stratified sampling.
That is, when distributing samples to users it is ensured that each strata (for example, age range) gets each type of action (or treatment).
If one action is globally better for achieving the desired outcome $y^\ast$ than the others then the proportions of the collected positive response labels will reveal that.

A/B testing approach would give statistics on which action is globally better.
However, it could be that for a particular group of people yellow envelopes lead to a better response rate,
while for other group blue envelopes lead to a better response rate.
A/B testing itself would not capture such subgroups which react to actions in different ways.
We are particularly interested in identifying and learning such local dependencies with the actionable attribute.
The following toy example illustrates the setting.



\begin{figure}
\centering
\includegraphics[width=0.2\textwidth]{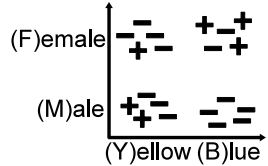}
\caption{A toy dataset in direct marketing: $x \in \{M,F\}$, $a \in \{Y,B\}$, and $y \in \{+,-\}$.}
\label{fig:example}
\end{figure}

Consider as a toy example the following situation in direct marketing.
Marketing offers can be sent either in blue (B) or in yellow (Y) envelopes.
The color of an envelope is the actionable attribute.
Suppose offers are sent to 20 persons (10 male (M) and 10 female (F)) choosing the envelope color at random.
Suppose the marketing manager gets \emph{six} positive responses $P(+)=0.3$, three from the blue envelopes and three from the yellow envelopes.
Thus, she concludes that there is no globally better envelope, since $P(+|Y)=P(+|B)=0.3$.
However, there may be subgroups, which respond to the colors differently, as illustrated in Figure~\ref{fig:example},
where `+' means a positive reaction, `-' means no response.
From the figure we see that the females have responded better to the blue envelopes and the males have responded better to the yellow envelopes.
The following response probabilities have been observed: $P(+|F,Y)=0.2$, $P(+|F,B)=0.6$, $P(+|M,Y)=0.4$ and $P(+|M,B)=0$.

Thus, the marketer instead of sending envelopes at random could have sent blue envelopes to the female users and yellow envelopes to the male users.
In that case the expected number of positive responses would have increased from \emph{six} (random baseline) to \emph{ten} (intelligent selection) and the expected probability of positive response from this marketing camping would have increased from $P(+) = 0.3$ to $P(+) = 0.5$ as summarized in Table \ref{tab:example}, assuming equal number of males and females in the user database.
\begin{table}[h]
\centering
\caption{Expected positive responses $P(+)$ in the marketing example.}
\begin{tabular}{lccc}
  \hline
  strategy              & $P(+|M)$  & $P(+|F)$  & $P(+)$\\ \hline
  baseline random       & 0.2       & 0.4       & 0.3  \\
  simple all yellow     & 0.4       & 0.2       & 0.3  \\
  simple all blue       & 0         & 0.6       & 0.3  \\
  intelligent selection & 0.4       & 0.6       & {\bf 0.5}\\
  \hline
\end{tabular}
\label{tab:example}
\end{table}

Moreover, knowing the sensitivity of different subgroups the marketer can optimize the strategy further.
The probability is maximized if an offer is sent \emph{only} to females in blue envelopes, then $P(+)=0.6$.
Sending to selected (not all) individuals may be preferred if postage has high costs.

Can the marketer do even better than that in this situation? Likely yes, if she has access to additional information about the users. In such a case she could rank the historical users based on their sensitivity to actions and learn a model for predicting the correct actions focusing on the most sensitive users.
In the next section we introduce three approaches for constructing the action rules and a strategy for focusing the modeling process for achieving a better accuracy.


\section{Approaches for deciding upon actions}
\label{sec:method}


We present and analyze three heuristic approaches for finding action rules in a supervised learning manner.
The first approach simply builds a classifier and then uses it for finding the right action by trying each of actions and picking the one that corresponds to the prediction closest to the desired outcome.
The second approach reverses the supervised learning setting by adding the target label to the input space and considering action as a target attribute.
The third approach constructs a new training set treating the outcome as a context.



\subsection{The Seek approach}

This approach is naive in a sense that it does not learn the action rule ${\cal H}$ explicitly.
Instead, it learns a traditional classifier the mapping between an instance and the outcome ${\cal L}: (X a) \rightarrow y$, where each instance includes the actionable attribute $a$ such as it was recorded in the historical data.

Given an unseen instance different values of the actionable attribute are explored and the respective model outputs are observed.
The action, which leads to the output closest to the desired $y^\ast$ is selected. 
The approach is summarized in Algorithm~\ref{alg:wrap}.
\begin{algorithm}
\SetAlgoLined
\SetKwInOut{Input}{input}\SetKwInOut{Output}{output}
\Input{labeled training data $({\bf X},{\bf a},{\bf y})$}
\Output{action rule ${\cal H}_{Seek}:X \rightarrow a$}
\SetKwFunction{distance}{distance}
\SetKwFunction{domain}{domain}
\BlankLine
Train ${\cal L}: (X a) \rightarrow y$ with $({\bf X},{\bf a},{\bf y})$\;
Decision making: given a new instance $X$\;
\For{$a_i \in$ \domain$(a)$}{output $\hat{y}_i={\cal}(X,a_i)$}
Decide ${\cal H}_{Seek}: a = a_j$, where $j = \arg \min_j$ \distance$(y^\ast,\hat{y}_j)$
\caption{The Seek approach}
\label{alg:wrap}
\end{algorithm}

The approach has the following limitations.
\begin{enumerate}
\item There is no explicit action rule; therefore, in practice the domain of possible actions should be limited to not too many alternative actions.
\item The maximum and the minimum values of $a$ need to be controlled in case $a$ is numerical.
\item Depending on the model class the predictive model ${\cal L}$ assumes a certain relationship between $a$ and $y$. For instance, linear models will learn either positive or negative relation between $a$ and $y$. In such a situation one type of action will always be better or always worse.
\end{enumerate}



We illustrate the approach with the previous toy example from Figure~\ref{fig:example}.
The dataset has four types of instances (a combination of male or female, and yellow or blue).
For each type we test whether different base classifiers can learn the true effects of actions from the data with several base classifiers.
Table \ref{tab:method1} presents the results, where the desired outcome is $P(+|F,Y)<P(+|F,B)$ and $P(+|M,Y)>P(+|M,B)$, the classifiers that satisfy it are marked in bold.
\begin{table} [h]
\centering
\caption{Positive response rates using the Seek approach on the marketing example.} 
\begin{tabular}{lcccc}
  \hline
                            & $P(+|F,Y)$    & $P(+|F,B)$    & $P(+|M,Y)$    & $P(+|M,B)$    \\ \hline
  TRUE probabilities        & 0.2           & 0.6           & 0.4           & 0  \\
  LEARNED probabilities:    &&&&\\
  Naive Bayes   & 0.39          & 0.39          & 0.21          & 0.21           \\
  logistic regression& 0.40         & 0.40          & 0.20          & 0.20           \\
  linear discriminant& 0.60       & 0.60          & 0.38          & 0.38           \\
  {\bf quadratic discriminant}& 0.27& {\bf 0.79}    & {\bf 0.63}    & 0.04 \\
  SVM (rbf)     & 0             & {\bf 1}       & 0             & 0         \\
  {\bf regression tree} & 0.2     & {\bf 0.6}     & {\bf 0.4}     & 0       \\
  kNN (k=5)     & 0             & {\bf 1}       & 0             & 0       \\
  {\bf neural network (4 hidden layers)}& 0.11    & {\bf 0.55}    & {\bf 0.66}    & 0     \\ \hline
\end{tabular}
\label{tab:method1}
\end{table}

It can be seen that linear classifiers do not make a distinction between the envelopes at all, since the true decision boundary is non linear, as it is a XOR type classification problem (see Figure~\ref{fig:example}, male and female groups with higher positive response are located on the opposite corners).
The non linear crisp-output classifiers (SVM, kNN) correctly identify the proffered color for females, but not males, since the negative response is dominated among males. The regression tree correctly identifies the posterior probabilities, the quadratic discriminant and the neural network give approximately correct probabilities in terms of the desired outcome.

\subsection{The Twist approach}

This approach adds the target label to the attribute space and treats the actionable attribute as a label (i.e.\ as the target attribute to predict).
The task now is to learn the following mapping ${\cal H}_{twist}:(X,y) \rightarrow a$.

For an unseen instance the label $y$ is not known at the prediction time when we need to augment the input space.
Instead, we can add the desired value of the label $y^\ast$.
The approach is summarized in Algorithm~\ref{alg:twist}.
\begin{algorithm}
\SetAlgoLined
\SetKwInOut{Input}{input}\SetKwInOut{Output}{output}
\Input{labeled training data $({\bf X},{\bf a},{\bf y})$}
\Output{action rule ${\cal H}_{twist}:X \rightarrow a$}
\BlankLine
Train ${\cal H}_{twist}: (X y) \rightarrow a$ with $({\bf X},{\bf a},{\bf y})$\;
Decision making: given a new instance $X$ decide ${\cal H}_{twist}: (X y^\ast) \rightarrow a$, where $y^\ast$ is the desired outcome.
\caption{The Twist approach}
\label{alg:twist}
\end{algorithm}


When augmenting the input space 
the approach assumes that for every instance $X$ the desired outcome $y=y^\ast$ is possible, which may not necessary be always true.
However, that is not a problem as long as the other outcomes ($\neg y^\ast$) are equally ineffective, or the classification problem is binary.

We illustrate the approach with the previous toy example using different base classifiers.
The results are presented in Table \ref{tab:method2}.
The desired outcome is $P(Y|F,+)<P(B|F,+)$ and $P(Y|M,+)>P(B|M,+)$, the methods that satisfy it are marked in bold.
As in the previous case, the linear classifiers are not able to capture non the linear subgroups subgroups, the non linear classifier perform well on this task.
\begin{table}[h]
\centering
\caption{Positive response rates using the Twist approach on the marketing example.}
\begin{tabular}{lcccc}
  \hline
  classifier    & $P(Y|F,+)$    & $P(B|F,+)$    & $P(Y|M,+)$    & $P(B|M,+)$    \\ \hline
  Naive Bayes   & 0.5           & 0.5           & 0.5           & 0.5           \\
  logistic regression& 0.5          & 0.5           & 0.5           & 0.5           \\
  linear discriminant& 0.5        & 0.5           & 0.5           & 0.5           \\
  {\bf quadratic  discriminant}& 0.27& {\bf 0.75}    & {\bf 0.65}    & 0.06  \\
  SVM (rbf)     & 0             & {\bf 1}       & 0             & 0         \\
  {\bf regression tree} & 0.33    & {\bf 0.75}    & {\bf 0.63}    & 0       \\
  kNN (k=5)     & 0             & {\bf 1}       & 0             & 0       \\
  neural network (4 hidden layers)& 0.25          & {\bf 0.74}    & 0.50          & 0.63    \\ \hline
  TRUE          & 0.25          & 0.75          & 1             & 0  \\
  \hline
\end{tabular}
\label{tab:method2}
\end{table}

\subsection{The Contextual approach}


This approach considers the outcome as a hidden context \cite{Harries98} when modeling the actions.
More specifically, in this approach the outcome is not part of the input attributes, but rather it describes an implicit group of similar users.
The users who responded positively (`+') form one context.
Those who did not respond (`--') form the other context.


We assign each instance into the desired context by manipulating the value of the actionable attribute.
We construct a new training set for learning the action rule ${\cal H}_{context}$.
In this training set we define labels in a specific way. Given a binary actionable attribute we make the following assumptions:
\begin{itemize}
\item users responded positively (+) either because they were certain about their positive attitude independently of the envelope, or because the \emph{right} color of envelope was sent;
\item users responded negatively (--) either because they were certain about their negative attitude independently of the envelope, or because of a \emph{wrong} color was sent.
\end{itemize}
Thus, to all positive users (+) we assign the labels according to the original colors of the envelopes, which were sent to them.
To the negative users (--) we assign the \emph{opposite} colors as they originally received, hoping that the other color might have changed their decision.

Using the new training set $({\bf X},{\bf \bar{a}})$ constructed as described, we build a classifier ${\cal H}_{context}: X \rightarrow a$.
The approach is summarized in Algorithm~\ref{alg:context}.
\begin{algorithm}
\SetAlgoLined
\SetKwInOut{Input}{input}\SetKwInOut{Output}{output}
\Input{labeled training data $({\bf X},{\bf a},{\bf y})$}
\Output{action rule ${\cal H}_{context}:X \rightarrow a$}
\BlankLine
Construct a modified training set $({\bf X},{\bf \bar{a}})$, where
\For{each training instance $X_i$}
{
\eIf{$y_i=\text{'+'}$}{$\bar{a}_i:=a_i$}{$\bar{a}_i:=\neg a_i$}
}
Train ${\cal H}_{context}: X  \rightarrow a$ with $({\bf X},{\bf \bar{a}})$\;
Decision making: given a new instance $X$ decide ${\cal H}_{context}: X \rightarrow a$
\caption{The Contextual approach}
\label{alg:context}
\end{algorithm}



This approach is restricted to binary actionable attributes, but not limited to binary outcomes.
One more potential limitation of this approach is that by modifying the training set according to our assumptions we may introduce noise in the data.
The Contextual approach has an advantage over the other two.
If a learning problem is formulated in this way, the ground true labels for testing are in fact available.

We illustrate the approach with the previous toy example using different base classifiers.
Table \ref{tab:method3} presents the results.
The desired outcome is $P(Y|F)<P(B|F)$ and $P(Y|M)>P(B|M)$, the methods that satisfy it are marked in bold.
We see that all the classifiers except kNN do well in identifying the right action. 
\begin{table}
\centering
\caption{Positive response rates using the Contextual approach on the marketing example.}
\begin{tabular}{lcccc}
  \hline
  classifier    & $P(Y|F)$      & $P(B|F)$    & $P(Y|M)$    & $P(B|M)$    \\ \hline
  {\bf Naive Bayes} & 0.3       & {\bf 0.7}     & {\bf 0.7}     & 0.3           \\
  {\bf logistic regression}& 0.3    & {\bf 0.7}     & {\bf 0.7}     & 0.3           \\
  {\bf linear discriminant}& 0.3  & {\bf 0.7}     & {\bf 0.7}     & 0.3           \\
  {\bf quadratic discriminant}& 0.3 & {\bf 0.7}     & {\bf 0.7}     & 0.3           \\
  {\bf SVM (rbf)}     & 0       & {\bf 1}       & {\bf 1}       & 0         \\
  {\bf regression tree} & 0.3     & {\bf 0.7}     & {\bf 0.7}     & 0.3       \\
  kNN (k=5)     & 0             & {\bf 1}       & 0             & 1      \\
  {\bf neural network (4 hidden leayers)}& 0.17    & {\bf 0.83}    & {\bf 0.83}    & 0.17    \\ \hline
  TRUE          & 0.3           & 0.7           & 0.7           & 0.3  \\
  \hline
\end{tabular}
\label{tab:method3}
\end{table}

\subsection{Focusing the learning process on the sensitive cases}

\begin{figure}
\centering
\includegraphics[width=0.7\textwidth]{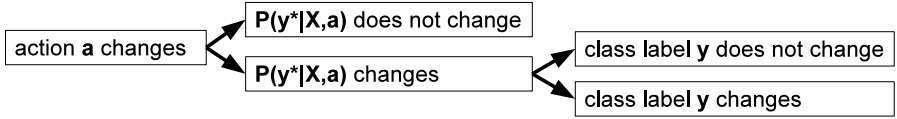}
\caption{Possible outcomes of changing the actionable attribute.}
\label{fig:possible}
\end{figure}

We presented the approaches in a simplified scenario, which assumes that the contribution of an actionable attribute to the label is uniform for all instances.
In more complex real world data the impact of an action may be different for every instance.
Changing the action may have several possible outcomes to the class label as illustrated in Figure~\ref{fig:possible}.
In the classification setting we are primarily interested in identifying the correct actions for those instances, for which changing the action changes the class label. Thus, we aim at focusing the modeling process on those instances.

\begin{figure}
\centering
\includegraphics[width=0.4\textwidth]{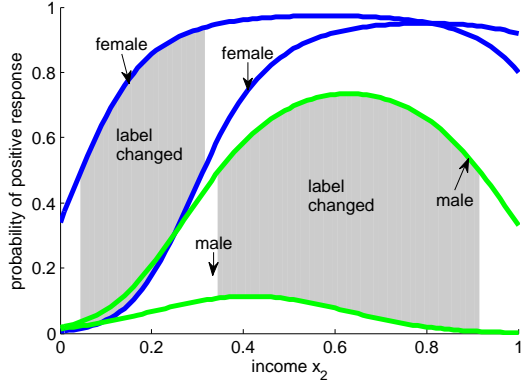}
\caption{Responses to actions on the toy dataset.}
\label{fig:toyresults1}
\end{figure}

We analyze these possible outcomes experimentally with an extended toy dataset shown in Table \ref{tab:toydata}.
Now there are two input attributes (gender and income), the same actionable attribute (envelope color) and the class label (response to the marketing letter).
We use the Seek approach with the quadratic discriminant classifier.
The resulting probabilities of a positive response are depicted in Figure~\ref{fig:toyresults1}.
\begin{table}[h]
\centering
\caption{The extended toy dataset.}
\begin{tabular}{ccccccccc}
  \hline
 $x_1$    & $x_2$    & $a$        & $y$      && $x_1$      & $x_2$    & $a$        & $y$    \\
gender    & income   & envelope   & response && gender     & income   & envepole   & response \\ \hline
         F&    0.00&    Y&        - &&     M&    0.55&    Y&       -\\
         F&    0.05&    Y&        - &&     M&    0.65&    Y&       +\\
         F&    0.20&    Y&        - &&     M&    0.85&    Y&       +\\
         F&    0.30&    Y&        - &&     M&    0.90&    Y&       -\\
         F&    0.35&    Y&        + &&     M&    0.95&    Y&       -\\
         F&    0.00&    B&        - &&     M&    0.60&    B&       -\\
         F&    0.10&    B&        + &     &M&    0.70&    B&       -\\
         F&    0.15&    B&        - &     &M&    0.75&    B&       -\\
         F&    0.25&    B&        + &     &M&    0.80&    B&       -\\
         F&    0.40&    B&        + &     &M&    1.00&    B&       -\\
         \hline
\end{tabular}
\label{tab:toydata}
\end{table}

In the earlier toy example (Section \ref{sec:background}) the probability of positive response was maximized ($0.6$) when sending the offer \emph{only} to females in a blue envelopes.
In the present example if a marketer sends any envelope for a female with the income over $0.25$,
the expected probability of positive response becomes much higher ($P(+|F, x_2>0.25) \approx 1$).
Moreover, sending yellow envelopes to the males with income around $0.75$ also gives a higher probability of response ($P(+|M, x_2>0.75) > 0.6$).

As before, the blue envelopes are more effective for females and the yellow envelopes are more effective for males.
However, we observe that there are certain situations (highlighted in \emph{grey} in the figure) in which the response (class label) changes as a result of change in the envelope color (action), while in the rest of the cases the response remains the same.
To achieve the best performance we aim at focusing the training procedure in such a way that the predictive model would specialize in the sensitive cases.
\emph{Sensitive cases} are the instances for which the outcome changes as a result of an alternative action.
As we will see, the sensitive cases are not necessarily the same as the instances close to the decision boundary.
Note that even if the predictions in the remaining cases are inaccurate, they will not affect the overall positive response rate, because they are not sensitive enough to the action.

\begin{figure}
\centering
\includegraphics[width=0.7\textwidth]{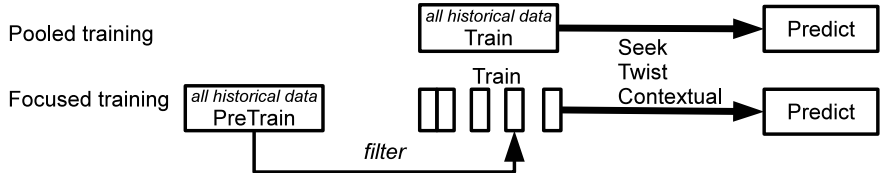}
\caption{The pooled and the focused training procedures.}
\label{fig:pretrain}
\end{figure}

Figure~\ref{fig:pretrain} illustrates the \emph{focused} training procedure as compared to training a model on all the historical data (which we refer to as \emph{pooled} training). 
The focused training divides the historical data into into pre-training and training subsets and uses two levels of training.
Firstly a filter is trained on pre-training data. The filter is a classifier ${\cal L}: (Xa) \rightarrow y$.
Next this filter is applied to the training dataset trying all the values of the actionable attribute and inspecting the resulting output of the filter.
The instances for which the output changes as a result of changing the action are selected. 
These selected instances then are used to train the action rule (employing one of the proposed approaches: Seek, Twist or Contextual).

\subsection{Applicability of the approaches}

The proposed approaches solve the problem of maximizing the probability of the desired outcome $P(y^\star|X,a)$ for any user $X$
by producing a model ${\cal H}:X \rightarrow a$ for determining the optimal action for a new user.
The model is learned and optimized for suggesting an action such that the probability of the desired outcome is maximized.

The three approaches for deciding upon an action make different underlying assumptions, as summarized in Table~\ref{tab:3app}.
\begin{table}[h]
\centering
\caption{Summary of the proposed approaches for choosing an action.}
\begin{tabular}{lcccc}
  \hline
            		& type      		& $a$           	& $y$      & labels for testing  \\ \hline
  Seek      	& wrapper   	& categorical   	& any      & no \\
  Twist     		& filter   		& any           	& any      & no  \\
  Contextual	& filter    		& binary        	& any      & yes  \\
  \hline
\end{tabular}
\label{tab:3app}
\end{table}

The proposed focused training procedure is intended to optimize the predictive performance in the situations where the correct prediction of which action to take affects in final the outcome (the response of the user).

One could argue that the action rules obtained using the proposed approaches may not necessarily capture causal relationships, but rather be a result of correlations due to, for instance, sampling bias. In order to make sure that the action rules are causal and thus actionable, we need to make sure that the actions in the historical datasets that we use are resulting from randomized trials rather than a deterministic procedure~\cite{Rubin78}. In other words, we need to make sure that $X$ and $Pa$ are independent.
Randomized trials are very common in medicine, economics, psychology, computational advertising and other domains for ensuring that the discovered relationships are causal.
In the considered toy envelope example it would mean that the colors of envelopes need to be assigned to the users at random.

Nevertheless, our approach is helpful for finding (not necessary causal) relationships also from field experiments, i.e.\ for performing secondary data analysis rather than mining purposefully collected data through the randomized trials. 
In these cases to confirm that found relationships are valid and useful for adaptation at the individual level, the company needs to construct a randomized trial and perform the hypothesis testing. However, in practice, it is useful for companies to see the potential of such fine-grained personalization before going for deployment and A/B testing. Our approach may give a valuable information to a domain expert to decide whether the potential benefit worths going for A/B testing and redesign if the objectives are met.


\section{Experimental evaluation}
\label{sec:experiments}


This section presents experimental evaluation of the proposed approaches.
We start with discussing the evaluation challenges.
Then we test the approaches using synthetic data.
Finally, we analyze the performance of the proposed approaches with two real-world case studies in web analytics and e-learning domains.

\subsection{Evaluation challenges and approach}
\label{sec:evaluation}

Evaluation of learning with actionable attributes on real data is challenging,
since the outcomes of alternative actions applied to the same user are not known.
Therefore it is not possible to test the \emph{what-if} scenario and to measure directly whether the suggested choice of an action is correct.

In uplift modeling the accuracy of a ranker is approximated with gain curves \cite{Radclife07}, which is not suitable for our setting as it requires one action to be globally better than another action and assesses the ranking of instances produced by a classifier rather than a selection from alternative actions.

In the alternative action setting 
a produced action rule ${\cal H}:X\rightarrow \hat{a}$ suggests an optimal action for that user, to maximize the posterior probability of the desired outcome ($y^\ast$).
Four situations are possible as a result:
\begin{enumerate}
\item $y=y^\ast$ (positive instance) and $a=\hat a$ (the suggested value of the actionable attribute is the same as the original one);
\item $y \neq y^\ast$ and $a=\hat a$;
\item $y = y^\ast$ and $a \neq \hat a$;
\item $y \neq y^\ast$ and $a \neq \hat a$.
\end{enumerate}
The first case can be counted as correct. The other three cases are not trivial to interpret.
In the second case we do not know if $y=y^\ast$ is possible at all for the given instance.
If it is possible, then the model made a mistake. Otherwise, it is not a mistake.
In the third case we do not know if $y=y^\ast$ is possible with other value of the actionable attribute ($\neg a$). If it is not possible, then this is a mistake, otherwise it is not.
In the fourth case it is not clear if the suggested action have changed the outcome or not, but it is not a mistake anyway.

To overcome these challenges 
a live online testing would be needed, which is out of the scope of this study.
Therefore we first test the proposed techniques on synthetic datasets where we know the ground truth distribution and thus the outcomes of the alternative actions.
Then we conduct two case studies with real data where we concentrate on exploratory analysis of the experimental
results to identify the effects of focused learning.

\subsection{Experiments with synthetic data}

The goal of these experiments is to analyze the performance of the proposed approaches and assess whether the proposed focused training helps to improve the performance.
We generate two datasets using the following procedures.

\subsubsection{Dataset generation}

\begin{figure}
\centering
\begin{tabular}{cc}
\includegraphics[width=0.2\columnwidth]{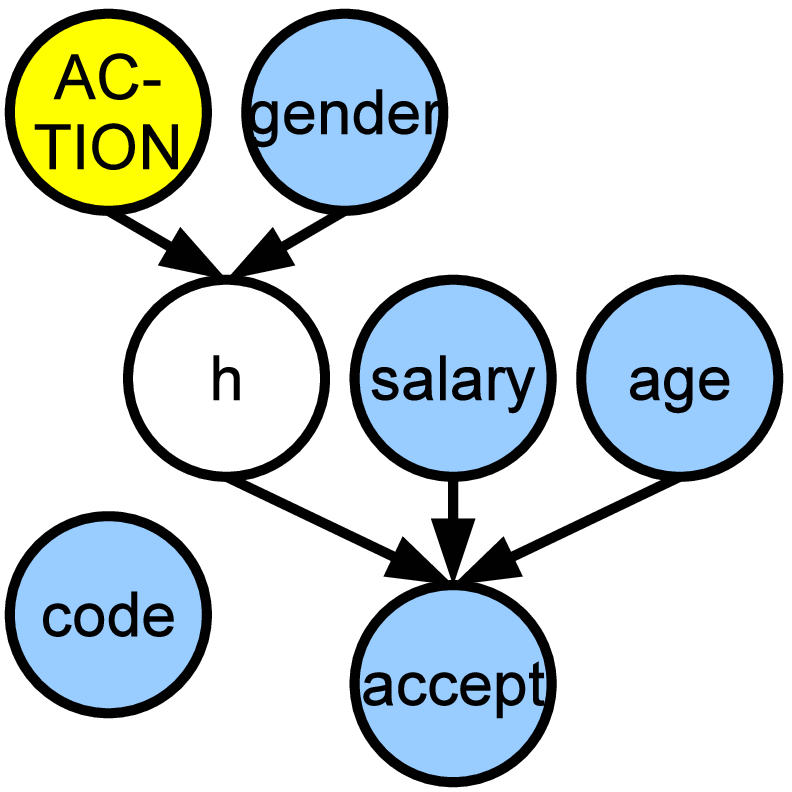} &
\includegraphics[width=0.2\columnwidth]{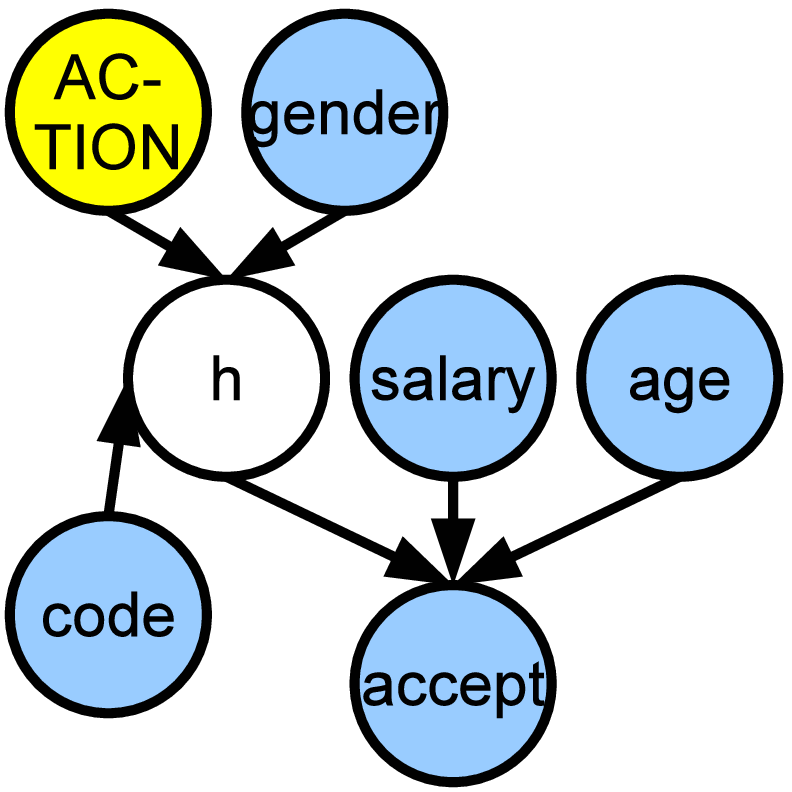} \\
(a) Data I & (b) Data II
\end{tabular}
\caption{Synthetic datasets, shaded circles indicate observed variables.}
\label{fig:data1}
\end{figure}

Our first dataset is designed around the marketing example discussed earlier
and corresponds to the belief network illustrated in Figure~\ref{fig:data1}~(a).
Suppose that a marketer sends an offer in a chosen form, i.e.\ \emph{color} of an envelop (a binary action, $a$) aiming to get a positive response ($y^*=`accept'$).
The response depends on two numeric variables, \emph{salary}, $x_1$ and \emph{age}, $x_2$. Besides, the unobserved preference ($h$) of an action,
which impacts the acceptance decision,  depends on a binary variable (\emph{gender}, $x_3$) and is independent of an observed categorical variable (\emph{postal code}, $x_4$).

\textbf{Data I} is generated with the following distributions:
$x_1 \sim u/2$,
$x_2 \sim u/2$, where $u \in [0,1]$ is a uniform random variable,
$x_3 \in \{`female',`male'\}$, $p(x_3=`female')=0.5$,
$x_4 \in \{1,2,3,4,5\}$, $p(x_4 = 1) = \ldots = p(x_4 = 5) = 0.2$.
The action $a \in \{0,1\}$ can be chosen by a decision maker.
The label is assigned $y = \rho\left( [x_1 + x_2 + h + \epsilon] \leq 0.5\right)$, where $\rho(\text{true}) = 1$, $\rho(\text{false}) = 0$, $\epsilon \sim {\cal N}(0,0.1)$ introduces random noise. The hidden preference $h$ depends on the \emph{action} and the \emph{gender} as presented in Table \ref{tab:pref}.
\begin{table}[h]
\centering
\caption{Hidden preference $h$ used for the synthetic data.}
\begin{tabular} {cc|cc}
$a$ &  $x_3=`female'$ & $x_3=`male'$ \\ \hline
`yellow'            & $-0.1$ & $0.1$ \\
`blue'              & $0.3$ & $-0.3$
\end{tabular}
\label{tab:pref}
\end{table}

In the dataset Data~I all the users are affected by the action.
We will need another dataset to assess the impact of focused learning, where some of the users would be indifferent to action, the impact will depend on the postal code variable $x_4$, as illustrated in Figure \ref{fig:data1} (b).

\textbf{Data II} is generated the same way as Data I except for the preferences. In this dataset preferences also depend on the postal code and within two postal codes of five the marketing action does not change the probability of accepting the offer.
Thus in Data~I $~57\%$ of instances change their labels as a result of action, while in Data~II only $~34\%$ change their labels.

\subsubsection{Experimental results with synthetic data}

We generate 10\,000 instances for training and a separate testing set of 100\,000 instances.
Since we know the data distribution, for the testing set we know what would be the label with the alternative action. Therefore we can assess what the \emph{relevant accuracy} of the classifier deciding upon action is.
The relevant accuracy is the accuracy of prediction among those testing instances for which the action $a$ changes the outcome $y$.
This measure is similar to \emph{recall} in information retrieval in a sense that we care in how many cases we chose the action correctly when it mattered. Measuring \emph{precision} with respect to all the instances in this case does not make sense, since the setting assumes that all the actions have equal costs and an action is performed anyway. For a user who does not react to an action any action is equally good, thus our evaluation criterion does not address that part of data.

\begin{table} 
\centering
\caption{Relevant accuracy (in $\%$) $\pm$ standard deviation over 100 runs.}
\begin{tabular} {llcccc}
\hline
        &           & Seek              & Twist             & Contextual        & Random baseline \\ \hline
Data I  & pooled    & $87.8 \pm 0.7$    & $91.1 \pm 0.7$    & $88.3 \pm 0.5$    & $50.0 \pm 0.2$ \\
        & focused   & $97.1 \pm 0.4$    & $97.2 \pm 0.6$    & $97.0 \pm 0.4$    & \\
\hline
Data II & pooled    & $79.1 \pm 4.2$    & $90.6 \pm 1.1$    & $88.4 \pm 0.7$    & $50.0 \pm 0.3$ \\
        & focused   & $96.6 \pm 0.8$    & $96.3 \pm 1.1$    & $96.5 \pm 0.7$    & \\
\hline
\end{tabular}
\label{tab:rezsin}
\end{table}


Table \ref{tab:rezsin} presents relevant accuracies achieved over 100 independent runs using the CART decision tree as the base classifier.
We observe that when using the pooled training on Data~I and Data~II the Twist approach performs the best with the Seek and Contextual approaches performing similarly.
All three strategies outperform the random baseline by a large margin.
The proposed focused training, which includes only a selection of instances in the training set, as expected, consistently outperforms the pooled training, which uses all the available for training.
Moreover, the relevant accuracy of the three techniques with the focused training are very similar to each other, suggesting that all three techniques are effective and accurate to decide upon action when equipped with the focused training.


\subsection{A case study in recommender selection}
\label{sec:masters}

MastersPortal.eu web portal provides information about Master study programmes in Europe. It was launched in 2007.
In March of 2010, the portal attracted nearly 1 million monthly visits and provided information on over 15 thousand programmes.
The revenue model for the portal was based on selling premium listings, web-based advertising campaigns and banner impressions.
Thus, it was vital to increase both the number of visitors and the number of pages viewed by each visitor.

Throughout the existence of the portal, the bounce rate of visitors was in the range of 55-70\%.
However, visitors referred by Google (that corresponds to half of the visits) had a much higher bounce-rate, about 90\%. 
They were referred directly to one of the programmes, and if not interested in that particular programme would go back to the Google's result page rather than exploring the portal. This was an unfortunate situation as likely the right information was available at one of the related pages of the portal.

In response to this situation, the portal helped visitors finding relevant information (and thus also promote staying longer within the portal) by building the recommender system. To decide whether a recommender would improve the situation and which recommendation approach would suit most, live testing with real users was conducted~\cite{Thijsthesis}. In testing the effect of using content-based and collaborative recommenders on the bounce rate was measured.
The results showed that with both either content-based or collaborative recommenders (but not a simple baseline recommender) it was possible to reduce the bounce-rate for visitors referred by Google from 90\% to 82\%. In the additional experiments it was studied whether factors such geographical location of users or topic of search had a (global) effect on preferring recommendations generated by collaborative filtering or content-based filtering approaches. Some of the studied factors showed the statistically significant difference in the preferences.
We use the data collected during this live testing for exploring the potential of learning with actionable attributes in determining for each user - recommendations generated with which approach would be preferred. The actions for the users were assigned at stratified random.

\emph{The goal of our experiments in this case study is to analyze how we can learn to choose the type of recommender (actionable attribute) in order to increase the expected rate of positive outcome, i.e.\ avoid visitors bouncing from MastersPortal.eu recommending them information that they need}.

\subsubsection{Dataset description}


The dataset contains information of 39477 \emph{human} visitors coming to MasterPortal.eu from Google during two weeks of live testing (March 17 - March 30, 2010).
Each visitor was directed into one of the two experimentation groups using a round-robin assignment approach. 
The experiment resulted in $48.3\%$ of visits shown content-based recommendations and $51.7\%$ collaborative.
The respective positive outcome rates (not bounced) were $17.6\%$ and $18.3\%$. Content-based recommendations performed slightly better, but the difference is not essential and not statistically significant.

We use the following 14 attributes for learning:
\emph{recommender type} as one actionable attribute, $a=\{collaborative, content\}$, 12 descriptive attributes $x_1$..$x_{12}$:
\emph{session count} (returning visitor or not),
\emph{continent} (Europe, Asia, ...), \emph{country}, \emph{screen resolution} (computer, phone, or gadget),
\emph{browser} (IE, Firefox or Chrome), \emph{referrer site}, six \emph{query related attributes} (query category inferred from query keywords), and the target attribute
\emph{outcome} (bounced or not, $y^*=$'not bounced' is the desired outcome).
The priors for the label are $17.9\%$ (not bounced) and $82.1\%$ (bounced).



We aim to focus the learning of an action rule to the borderline cases, the ones which are the most sensitive to the optimal choice of the recommender.
Note, that we do not use the information about the overlap in recommendations generated by content-based and collaborative approaches, while they might have provided the same recommendation in certain cases.



Experimental evaluation on MastersPortal.eu data 
consists of two parts.
In the first part we explore where the boundary cases in the data are.
In the second part, we test the proposed approaches and compare the focused learning technique with learning using all the training data.

\subsubsection{Where are the sensitive cases?}

The goal of the first experiment is to investigate the relation between the action of choosing the recommender type $a$ and the desired outcome of not bouncing $y^\star = +$.
Here we use the Seek approach, which allows to estimate numerically the probabilities of the desired outcome $P(+|X,a)$ with alternative actions.
We use the CART decision tree with continuous outputs the base classifier, which can be used as estimates of the posterior probabilities.

We split the data into equal sized train and test parts at random. We use the train part for training a predictive model ${\cal L}:X,a\rightarrow y$.
Then from the test part we prepare two alternative datasets:
one with collaborative recommender as an action $Test_A=({\bf X} {\bf a}_{collaborative})$ and another with content-based recommender $Test_B=({\bf X} {\bf a}_{content} )$.
We test the model ${\cal L}$ on both sets separately and analyze the resulting posterior probabilities.

To support the validity of the proposed approaches, we are interested to find from the data:
(1) whether we can find any groups of users that are sensitive to actions in different ways, and (2) whether the users that are the most sensitive to actions can be found anywhere in the instance space.
A positive answer to question (1) would indicate a need for our approaches, and question (2) would indicate a potential for the proposed focused training procedure.



\begin{figure}[t] 
\centering
\includegraphics[width=0.4\textwidth]{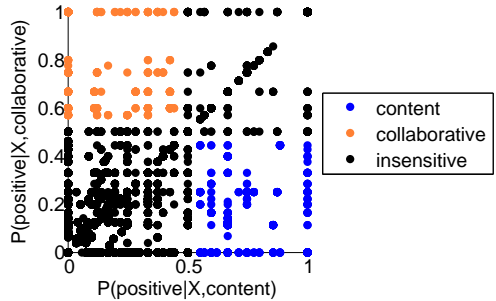}
\caption{Sensitivity to changing the action in the web analytics case study.}
\label{fig:delta}
\end{figure}

Figure~\ref{fig:delta} plots the estimated probabilities of positive outcome given one action (content based recommender) against the posterior probabilities given the alternative action (collaborative). The plots confirm our intuition discussed above.
We see that there are at least three groups of users (encoded by different colors): the users that are likely to  stay on the site given the content type recommender, but leave given collaborative (blue), the users that would stay given the collaborative recommender (orange) and the users that are indifferent (black).
More importantly, we see that the users that are sensitive (blue and orange) are not concentrated around the classification decision boundary (e.g. $P(+|X,a)=0.5$), but can be found all over the instance space. It means that a user that was very unlikely to be positive may become very positive as a result of the right action, but on the other hand, another user may be completely insensitive to action.
In total $11\%$ of the users  are likely to change their mind towards positive outcome given the correct action.
We aim to focus the training process on such users.




This analysis confirms that sensitivity to action is not the same as closeness to the decision boundary.
The actionable attribute may give a large impact to the posterior even being far from the decision boundary and the other way around.
We aim at focusing the process of action rules learning to the cases when the value actionable attribute makes high impact to the final outcome. 

\subsubsection{Learning focused on the sensitive cases}

The goal of the second experiment is to compare the behavior of the three proposed approaches in two training setting; with the focused training versus the pooled training.
With the real-world case study we cannot verify whether the action was correct, as the ground truth is not available for us.
Nevertheless, we can demonstrate, that the proposed approaches allow to distinguish between two actions for groups of similar users.
Hence, for analyzing the performance we group the testing instances into six groups based on the continents of origin of the site visits.
We inspect how distinct the recommendations for an optimal action are within each group.
We expect to see different actions recommended for different groups; in this particular case we expect different preferences and behavior of the users coming from different continents. Of course, it is very crude to assume that all people from the same continent would have similar preferences.
However, this allows to approximately analyze the performance when the ground truth is not available to us.

The experimental protocol is as follows.
We divide the data at random into three parts of equal size: pre-training, training and testing.
We use the pre-training set to build a filter for filtering training data for the \emph{focused} learning.
Recall that the filter identifies the instances which change the label as a result of change in action, the way it was within the grey areas in Figure~\ref{fig:toyresults1}.
We use the CART regression tree as a filter.
We present the results averaged over 100 runs with independent splits of data into training and testing.

\begin{figure}
\centering
\begin{tabular}{cc}
  \includegraphics[width=0.3\textwidth]{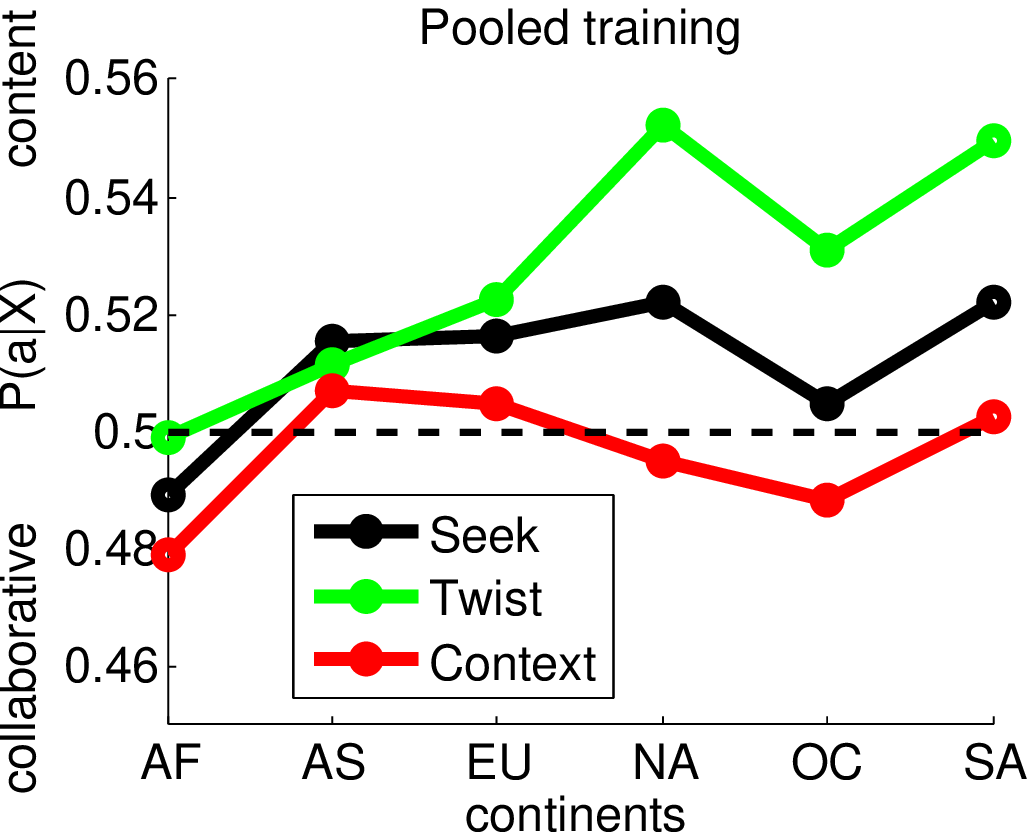} &
  \includegraphics[width=0.3\textwidth]{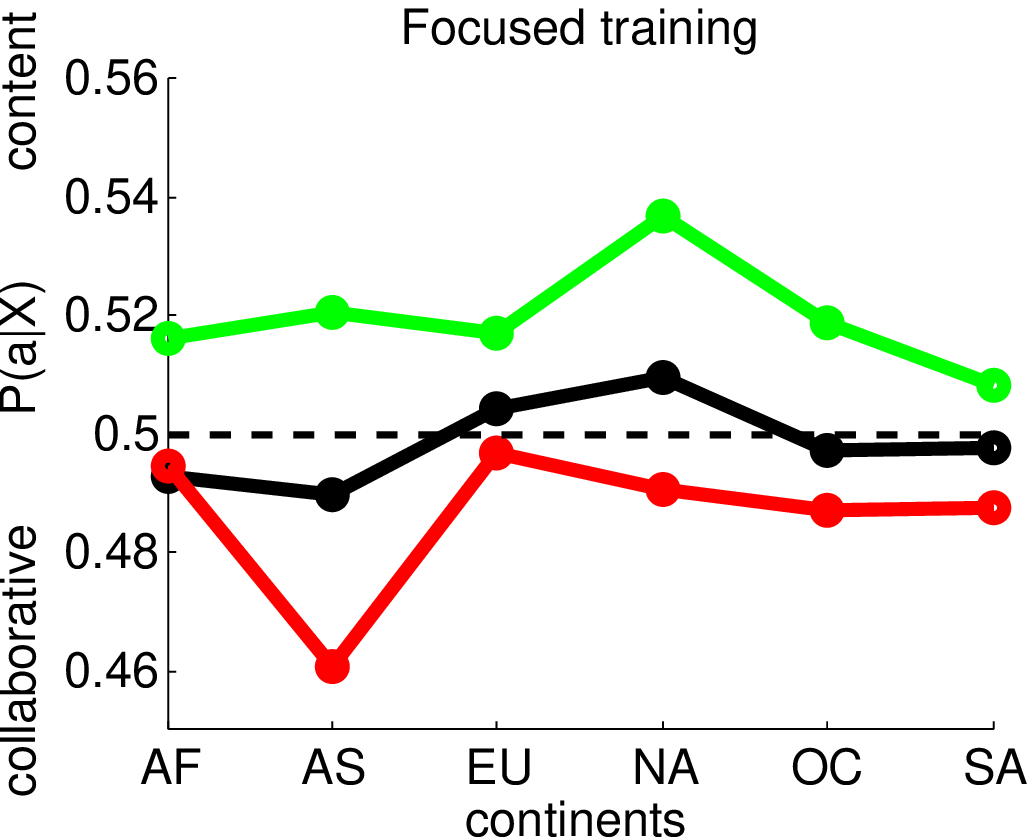}
\end{tabular}
\caption{Recommended actions with pooled and focused training in the recommender selection case study.} 
\label{fig:context}
\end{figure}

Figure~\ref{fig:context} presents the average probabilities of the proposed actions with the pooled training and the focused training.
We analyze whether the recommendations for actions are consistent within continents.
If we observed a flat line close to $0.5$, that would mean that the model was indifferent and thus could be suspected to be not effective.

We see that all three approaches more or less agree on the suggested actions.
For instance, a collaborative recommendation is more often suggested for the users from Africa, while a content based recommender is more often suggested for the users from Asia. In fact, these results are consistent with the findings of a study using A/B testing \cite{Thijsthesis}, shown in Table \ref{tab:africa}.
\begin{table}[h]
\caption{Results using A/B testing presented in \cite{Thijsthesis}.}
\centering
\begin{tabular}{lc}
\hline
User group & Preferred action \\
\hline
Africa & Collaborative\\
Asia & Content\\
Europe & Content\\
North America & Undecided\\
Oceania & Undecided\\
South America & Undecided\\
\hline
\end{tabular}
\label{tab:africa}
\end{table}

We observe that the pooled training leads to recommending the content-based approach in majority of the situations, while the focused approach nicely balances between the two alternative actions.
Aligning the suggestions produced by the proposed approaches with these results we can conclude that in the pooled training the recommendations by all three approaches are reasonable. However, in the focused approach the Seek and the Contextual approaches revert the suggestion for the visitors from Asia from content to collaborative.
On the contrary, the Twist approach shows a consistent behavior and has a clearly distinctive pattern.
Hence, we would advocate to follow the suggestions by the Twist approach.

\subsection{A case study in feedback personalization}
\label{sec:tue}

The second case study with real data relates to the effect of feedback in the online assessment of students knowledge and skills.
Students are answering multiple choice partial exam questions. They are requested to provide an answer and also indicate how certain they are regarding this answer.
After answering a question a student can request example-based or theory-based feedback, or in some cases the system can also decide to give such feedback automatically. 
The students are motivated to take it seriously, because related questions might still appear during the same exam, thus it is valuable to understand the right answer not just out of a curiosity but also because this could help to score better during the exam.
It is expected that different students might benefit more or less from one or the other type of feedback depending on their knowledge and their learning style.

\emph{The goal of our experiments in this case study is to analyze how we can learn to choose the type of feedback (actionable attribute) to maximize the probability of a correct answer (desired outcome) for the similar related questions present in the exam}.

\subsubsection{Dataset description}

The data was collected as a part of the PhD study~\cite{Vasilyevathesis} investigating the problem of feedback tailoring in e-learning setting, particularly during the examination. 

At Eindhoven University of Technology (TUE) during the spring term of 2007-2008 there were 73 students participating in the partial exams for the Human-Computer Interaction (HCI) course.
Each student was asked 11 exam questions, some of which were related to each other; i.e.\ if a student understands how to answer A, (s)he can also answer a related question B. Designed feedback for question A is aimed to help students to understand it better and thus increases chances to answer B correctly.

The automated assessment system could decide to show the certain type of feedback directly or recommend it for examining. The decisions of the system were based on the predefined personalization rules taking into account the correctness and certitude of the students response and his or her learning style. It was expected that in most of the cases each student voluntarily or forcibly would be exposed to one of the two existing explanations -- example-based and theory based. However, it was possible that in some cases the students did not see any feedback at all (e.g.\ if they answer correctly with high confidence and are not interested to request the feedback) and in some other cases the students would study both types of feedback (e.g.\ if the students after studying one type of feedback opted also for the other type believing that studying that not yet seen type of feedback would be still helpful).
For our experiment we filtered out the instances cases when no feedback was shown and cases when both types of feedback were shown to/requested by a student for a particular question. 
After filtering our dataset contains information of 571 instances; on average 8 responses per student. 
Example-based explanations have been seen by students in $29\%$ of the cases, and theory-based explanations -- in $71\%$ of the cases.
The respective positive outcome rates (correct answer) were $59\%$ and $35\%$.
Example-based feedback was used only one third of times, but lead to twice better related question correct answering ratio than for the theory-based feedback. 

We use the following $13$ attributes for learning:
one actionable attribute \emph{feedback}, $a=\{exmaple,theory\}$ -- example-based or theory-based type of feedback to show,
$11$ descriptive attributes $x_1$..$x_{11}$ including
\emph{first question} (categorical, referrer),
\emph{correctness of the answer} (binary, correct or wrong),
\emph{certainty of the answer} (binary, certain or not),
\emph{feedback} (binary, shown or requested), 
\emph{answer time} (binary, read or only seen/skimmed),
\emph{second question} (categorical, referrer), and
$5$ attributes determining \emph{learning styles}; and the last one is the \emph{outcome}, $y=\{correct,wrong\}$ -- correctness of the answer on the related question, $y^*=correct$ is the desired outcome.
The priors for the label are $42.2\%$ (correct) and $57.8\%$ (incorrect).

The task is to learn how to choose the optimal action (feedback type), so that the probability of the correct answer on the related question is maximized for a given student.


\subsubsection{Analysis of the results}

Similarly to the previous case study, we investigate how the change of a feedback type affects the expected probability of correctly answering a related question.
The experimental protocol is identical to the previous case study.
However, the TUE dataset has two distinctive features.
First, it is much smaller than the MastersPortal.eu dataset, therefore focused training approach typically uses less than 100 instances for training, the accuracy may notably suffer as a result of discarding training data due to small data size.
Second, in TUE data example-based feedback was on average twice more effective than theory-based, thus the action classifiers may gain bias towards recommending this type of feedback, while in MasterPortal.eu data the effect of both actions was balanced.
Nevertheless, we include the TUE case study primarily to support the main assumption behind this study that different individuals may prefer or may benefit from different actions.

\begin{figure} 
\centering
\includegraphics[width=0.4\textwidth]{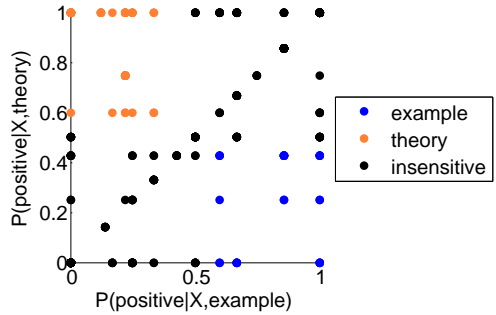}
\caption{Sensitivity to changing the action in the e-learning case study.}
\label{fig:deltaTUE}
\end{figure}

Figure~\ref{fig:deltaTUE} illustrates the effect of a change in action to the probability of a correct answer in the same manner as in the previous case study.
We again observe two groups of users that are sensitive to action and the group of users that are insensitive.
It means that a user that was very unlikely to be positive may become very positive as a result of the right action, but on the other hand, another user may be completely insensitive to action. There are $26\%$ of such users in this dataset.
The proposed approaches equipped with the focused training aim at identifying those users and deciding upon appropriate actions with them.

\begin{figure}
\centering
\begin{tabular}{cc}
\includegraphics[width=0.3\textwidth]{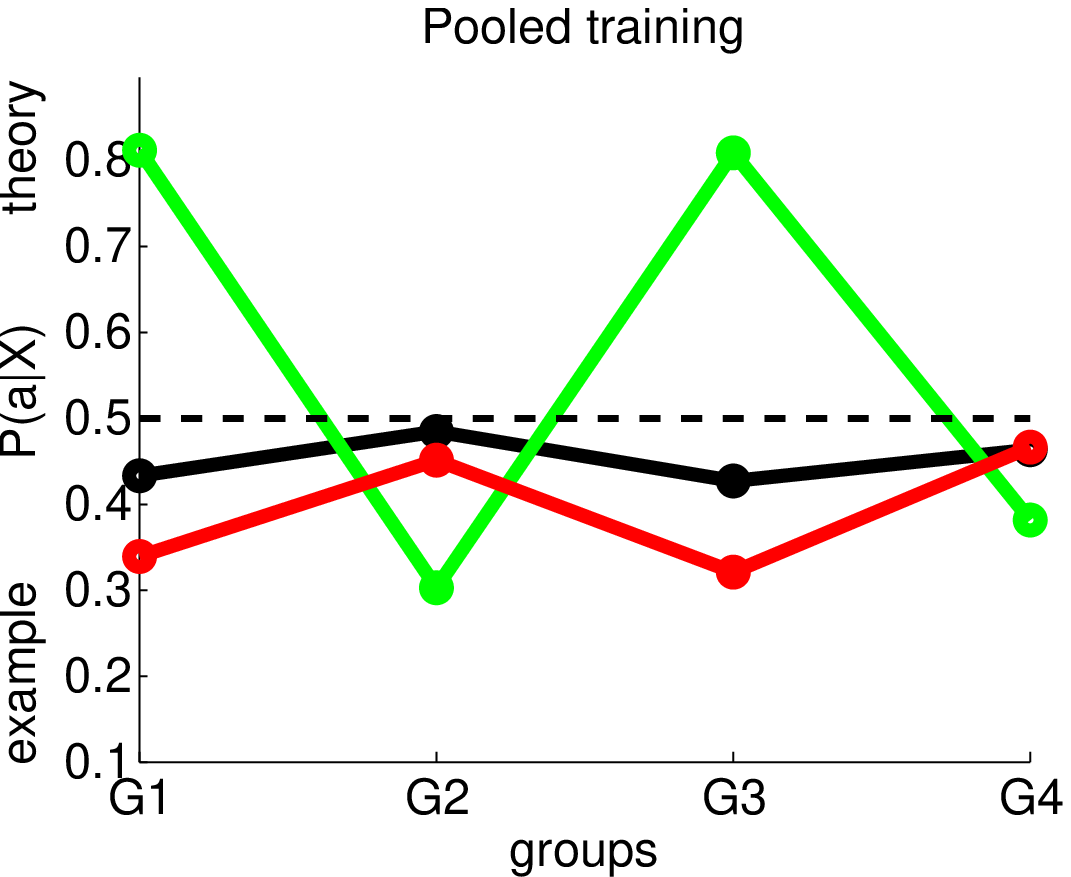} &
\includegraphics[width=0.3\textwidth]{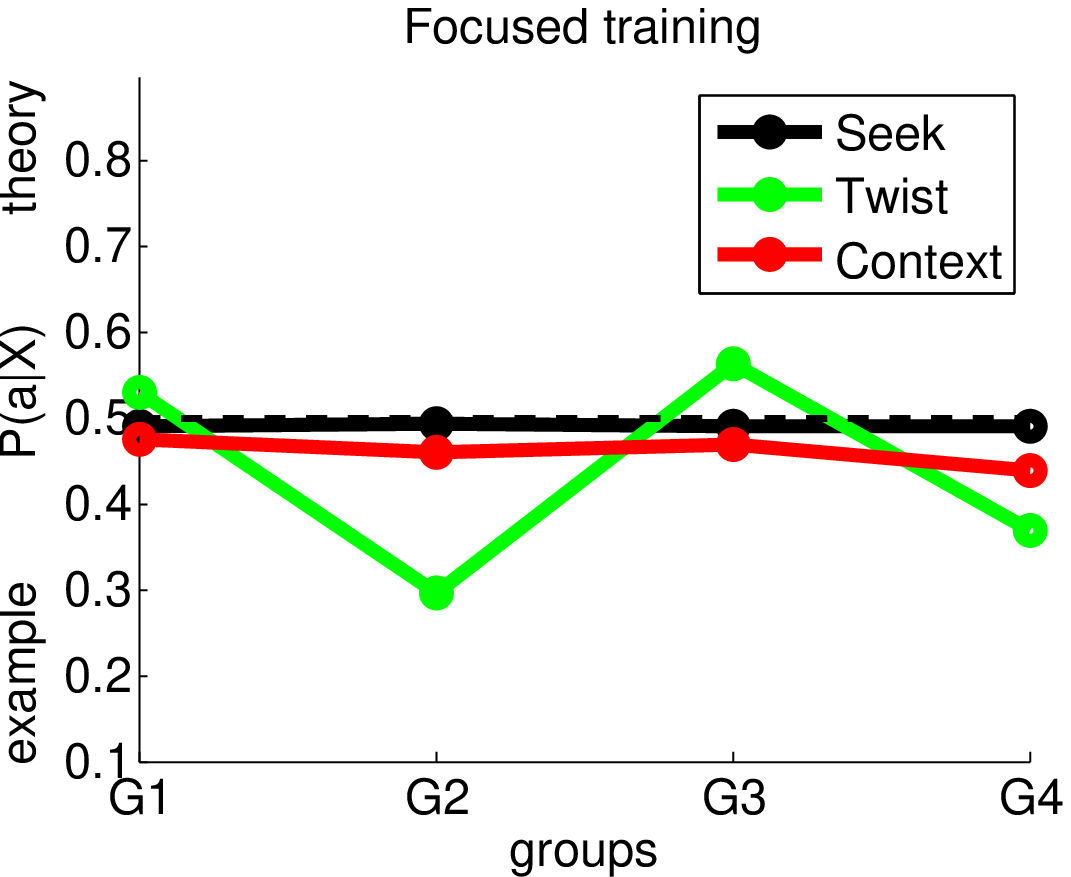}
\end{tabular}
\caption{Recommended actions with poole and focused training in the feedback personalization case study.}
\label{fig:context_tue}
\end{figure}

Finally, we test the three proposed approaches along with the pooled versus focused training, following the same protocol as for MasterPortal.eu.
We compare the results grouped by the certainty of the students answer and way the feedback was accessed.
In Figure~\ref{fig:context_tue} we analyze the effect of actions to the visitors within four groups: G1 includes the instances where a student indicated uncertainty of the answer, and got feedback automatically, G2 is for uncertain answers with a request for feedback, G3 is for certain answers with automatically shown feedback and G4 is for certain answers with a request for feedback.

We see that the Seek and Contextual approaches are not effective, they always prefer an example-based feedback (which is the artifact of the imbalanced dataset). The Twist approach, on the other hand, clearly demonstrates distinctive actions and can be considered the favorable approach in this case study. We see that theory appears to be a helpful form of feedback for the groups which did not explicitly request a feedback.
In case of the focused training the patterns are less distinctive; this performance may be attributed to the small overall size of the training set.
Nevertheless, the recommendations are consistent using the pooled and the focused training.

Overall, we would again advocate to follow the suggestions by the Twist approach.

\subsection{Implications and recommendations for practitioners}

The experiments with synthetic data where the ground truth is known demonstrate that the proposed approaches accurately identify the preferable actions, and that the focused training approach helps to increase the accuracy of actions.
The case studies confirm that different actions may be optimal for different instances and demonstrate the potential of the formulated set up and approaches.

We recommend using the Twist approach with the focused training procedure as a default option. It is the most generic with respect to its applicability to the different settings. Recall that it can deal with actionable attributes of different types, unlike Seek and Contextual approaches. Being a filter-type it is more efficient that Seek and unlike Contextual cannot introduce noise into the training data. Twist has shown more consistent behavior in our experiments. Unlike Seek, it is more robust to imbalance in global superiority of one action over the other. 

The choice of the classifier is not critical when the suggested focused training procedure is employed. But because of the pre-training and filtering is needed for focused training it require a larger training set to be available.

However, we would like to remind that from the decision support perspective, our approach is also useful for a domain expert because it provides an insight on the potential of the fine-grained personalization, i.e.\ the expected gain from applying adaptation at the instance level versus use of a globally best action.

The companies do not necessarily want to go for A/B testing or explore-exploit settings with every possible adaptation, because it may be costly to deploy and maintain while the potential benefit being not known.

Our approach allows the company to see what they can potentially achieve and help to decide whether to go for deploying into the operational settings and online testing of particular adaptive strategies or not.

We discussed the results of the conducted two case studies with the domain experts (data owners). In both cases they recognized the potential and concrete utility of our approach. First, they admired that our approach provided more fine grained capabilities for personalization than it would be feasible to do manually by an expert.
Second, in the case of Mastersportal, the experimental results were valuable to make a decision not to develop an adaptive system and even to keep and maintain only one type of the recommender system -- content-based recommender as this was considered to be more cost-effective than maintaining both recommenders and deploying and maintaining an adaptive mechanism based on the learnt model.
And in the case of TUE, we provided an additional confirmation to the domain expert that for the well defined situations the general feedback adaptation mechanism was already doing a good job and there was little room for making it more personalized, while for the most uncertain cases there was more room for fine-grained personalization. Were these insights and our techniques provided to the domain expert right after the data collection, her further research studies on tailoring feedback could be planned more effectively than they actually were, helping to avoid testing of less promising feedback tailoring strategies.


\section{Related work}
\label{sec:literature}

In this section we discuss related work in three groups.
Firstly, from the problem formulation point of view our study relates to discovering actionable knowledge or action rules.
Secondly, from the domain point of view the study relates to causality research.
Finally, from the technical point of view the proposed approach to focus training on the boundary cases relates to focused supervised learning.
The related work comes from different domains, including, but not limited to the action rule mining.
We are not aware of works directly related to the set up we explore in this paper.

A reader is refereed to recent surveys on mining actionable knowledge \cite{He05,Cao12}, which discusses data mining for decision support (follow up actions).
The role of domain experts is emphasized in producing action rules from data mining results.
The implications of data mining to the actions have been under discussion for more than a decade \cite{Adomavicius97}; however, these investigations typically focus on different aspects of knowledge discovery process than our study. In particular, actions are not learned directly from the data, but follow up actions as a result of discovering an interesting rule are considered. Our learning process is focused on learning action rules from data.
Discovering action rules either requires prior extraction of classification rules or the rules are mined directly from data.
The proposed Seek approach falls into the first type, while the Twist and the Contextual approaches represent the second type.

In relation to the desired outcome three mainstream paradigms for discovering action rules can be distinguished in the literature: association rules, rough sets and classification (decision trees).

Mining action rules without prior classification is often formulated as the association rule discovery problem \cite{He05b,Zhang08,Cao10,Su12}.
The main focus is how to formulate good interestingness measures to discover actionable knowledge: conditional lift (conditioned on the actionable attribute) \cite{Zhang08}, unified interestingness \cite{Cao10}, support and confidence with alternative actions \cite{He05b}.
The latter work also employs cost constraints for the actions.

A series of works employ the rough set techniques for mining action rules~\cite{Ras11,Ras_chapter,Ras00,Im08}.
The considered techniques are both with \cite{Ras00} and without \cite{Im08} prior classification rules.
In these works the action rules are the end goal for a decision making, the predictive analytics task is not explicitly associated.
An example in \cite{Ras_chapter} assumes a group of people who closed their bank accounts.
The goal is to find a cause why these accounts have been closed and formulate an action how to prevent it.

In \cite{Yang07,Ling02} actionable knowledge is extracted from \emph{decision trees}.
The approaches relate to the proposed Seek approach.
The idea is to inspect the resulting tree and find what attributes could be changed in order to end up in the decision leaf with higher probability of the desired outcome. Instead of assuming actionable and observable attributes, the authors introduce a cost matrix for changing the value of each attribute.
Impossible actions are marked with infinite costs (like changing from an adult to a child).

The uplift modeling \cite{Rzepakowski11,Hansotia02,Radcliffe11} is closely related with our setting.
In the uplift modeling two datasets are formed one with the action and another without the action.
An action is always considered to be desired, i.e.\ superior to no action, but the problem is that it incurs costs.
The task is to learn to quantify the profit of that action for a given example.
Although some uplift studies, such as \cite{Hansotia02} consider that actions may have negative effects on users,
our setting focuses on the situations in which there is no globally better action,
some actions work better for some individuals, other action for other individuals and there are individuals that are indifferent to actions.
Therefore we model it as a classification problem in a modified example space.

Both settings are sensible in different applications.
The uplift modeling setting is valid in treatment applications, e.g. in direct advertising when discounts are offered. In such applications every action has a significant cost and the question is to perform it or not for a given individual.
Our setting is of recommender nature rather than treatment.
It is relevant, for example, to web analytics, where actions (e.g. providing an example-based or a theory-based feedback in the web-based student assessment, or choosing the type of the recommendation approach) come at virtually no cost, thus we are interested to select the most appropriate one from a set of alternatives for each example at consideration. Note, that in our setting optimal action may be also no action.
Table \ref{tab:simon} summarizes the main difference of our setting with the recent study in uplift modeling \cite{Rzepakowski11}.
\begin{table}
\centering
\caption{Summary of the treatment (\cite{Rzepakowski11}) and the recommender settings.}
\begin{tabular}{lll}
\hline
Settings                    & Recommender           & Treatment\cite{Rzepakowski11} \\  \hline
costs of actions            & equal                 & asymmetric \\
effect of action to outcome & positive, no, negative& always positive \\
prediction                  & optimal action        & profit of action \\
control set (no action)     & not needed            & required \\
techniques                  & form training sets    & decision tree, Naive Bayes and regression specific criteria\\
applicable to               & any classifier        & two classifiers and a regression\\ \hline
\end{tabular}
\label{tab:simon}
\end{table}



The problem of discovering action rules is closely related to \emph{causality in feature selection} \cite{Guyon07}.
One of the earliest works related to choosing actions \cite{Geffner98} builds reasoning about actions from causality perspective.
A causality perspective can be considered complementary to our study in assessing the effects of actions.
In relation to causality, the effects of interventions to the decision rules are explored in \cite{Greco05}.
Causality~\cite{Aristotle} relates to exploratory data analysis when discovering the true cause and effect is the primary goal.
In our case we may potentially find rules which are linked by correlation, but not causality.
However, as we assume that the data distribution is stationary, we would not be able to perform an action on such rules that do not include causality, i.e.\ we would not be able to change the attribute value. Therefore, the domain experts would not indicate such attributes as actionable, as there would be no meaningful application task. For the further discussion of the variety of settings involving explanatory and predictive modeling we refer an interested reader to~\cite{Shmueli_EorP2010} and~\cite{Ebbes_Endogeneity}.

A probability based framework for value-change based actions is presented in~\cite{Chai04}
together with an approach for the Naive Bayes classifier as an instance of this framework.

From focused learning perspective there are remote relations to boosting~\cite{Freund95}, classification with reject option~\cite{Bartlett08} and evaluating classifier competence~\cite{Li06}. Boosting goes over a loop of training, each step putting more emphasis on learning the cases which were previously misclassified. Classification with reject option evaluates the regions of competence and is allowed not to output the decision if its confidence is too low.





\section{Discussion and Future Work}
\label{sec:discussion}

This study opens a potential for direct extensions and follow up studies for learning with actionable attributes, which we discuss in this section.

First, in this work we assumed that there is only one actionable attribute and that it is categorial.
Actionable attributes can also be ordered or real valued (e.g.\ a discount for the next purchase). In these cases a simple enumeration of actions and wrapper like approaches are not appropriate.

Existence of more than one actionable attribute brings more interesting challenges and opportunities. For example, the problem of recommender system itself can be formulated as learning with actionable attributes. Every user is described with a set of observed features (e.g.\ demographics and already rated items) and actionable attributes, i.e. items that can be recommended for this user. The task can be formulated as selecting the most promising actions such that the user would be interested or would like at least one of the recommendations.

Second, in this work we assumed that the costs associated with each value of the actionable attribute are the same or are zero. In some applications 
costs can be different for each action.
One straightforward approach to account for costs is to integrate them into the definition of the boundary cases. 

In general, if we deal with more than one actionable attribute, an optimization problem can be formulated: given a budget, decide on optimal strategy (including multiple actions) leading to the highest profit.

Another interesting addition is related to the fact that costs of an action is not necessary the same for all the examples (e.g.\ the price of sending an envelope to remote areas).

Besides costs, additional constrains can be introduced.
Some actions might be impossible or not allowed to perform together, while others can be complementary.
Impossible or not allowed actions or combination of actions can be simply marked with infinite costs, e.g.\ giving simultaneously two different medications, which cannot be used together. 

Third, in this work we assumed that there is no direct relation between the action and the target (e.g.\ the color of an envelope does not increase the value or reduces the price of the product). However, if we consider a mortgage application or placing a product `on sale' in a supermarket the action contributes to the outcome directly.
This assumption potentially can be incorporated in the learning process.

Cost-sensitive supervised learning, in which obtaining the value of each attribute for an user is associated with a particular cost (e.g.\ the cost of a medical test), may be also seen as close setup.
Indeed, a doctor may decide which test to perform to better diagnose a particular patient. The type of a diagnostic test or its outcome is not expected to cure a patient directly, i.e.\ change the label from the undesired to desired; the doctor is interested in the accuracy of the diagnosis. But the doctor also needs to decide on the type of treatment for the patient and the choice may effect the desired outcome of curing the patient. Thus, these are two interesting variations possible with our problem formulation, the first would be close to the utility-based or cost-sensitive classification setup, while the second share commonalities with uplift modeling.

All these variations lead to the formulation of interesting optimization problems, subject to our further research.


\section{Conclusion}
\label{sec:conclusion}

The paper defined a set up of focused supervised learning with actionable attributes and further research directions.
This problem has been studied from different perspectives in different areas of data mining, our study presents a novel viewpoint motivated by predictive analytics domain. We emphasized (1)~a classification setup where there is no globally better action, different actions may be preferable for different individuals, (2)~focused learning, motivated by the notion that different instances are sensitive to the value of the sensitive attribute in different ways.

Our experimental analysis demonstrated that an accurate choice of action is essential for those users, which are on a borderline, and that this borderline is not limited to the classification decision boundary between positive and negative instances, i.e.\ it is not sufficient to reduce this supervised learning problem to predicting with a reject option.
The experiments on synthetic data  with the known ground truth demonstrated that the proposed approaches could accurately identify the preferable actions, and that the focused training approach helped to increase the accuracy of choosing the right actions.
We validated the performance of the three proposed approaches with the pooled or focused training using two real-world case studies in web user modeling domain.
The case studies confirmed that different actions may be optimal for different instances and demonstrated the potential of the proposed approaches.
Based on the results, we would advocate for using the Twist approach trained using the focused learning procedure.

\bibliographystyle{plain}
\bibliography{bibaction}

\end{document}